\documentclass{article}
\usepackage{iclr2021_conference,times}

\usepackage{amsmath}
\usepackage{algpseudocode}
\usepackage{algorithm}
\usepackage{graphicx}

\usepackage{url}
\usepackage{xcolor}
 \usepackage{authblk}

\newcommand{\eg}{{\em e.g.}}
\newcommand{\etal}{{\em et~al.}}
\newcommand{\ie}{{\em i.e.}}
\newcommand{\etc}{{\em etc.}}
\newcommand{\RNum}[1]{\uppercase\expandafter{\romannumeral #1\relax}}

\title{LINKS: A dataset of a hundred million planar linkage mechanisms for data-driven kinematic design}

\author[1]{Amin Heyrani Nobari \thanks{Address all correspondence to this author.}
}

\author[2]{Akash Srivastava}

\author[3]{Dan Gutfreund}

\author[4]{Faez Ahmed}

\affil[1]{
	Department of Mechanical Engineering\\
	Massachusetts Institute of Technology\\
	Cambridge, Massachusetts 02139\\
    Email: ahnobari@mit.edu}
    
\affil[2]{MIT-IBM Watson AI Lab\\
	IBM Research\\
	Cambridge, Massachusetts 02142\\
	Email: akash.srivastava@ibm.com
    }

\affil[3]{MIT-IBM Watson AI Lab\\
	IBM Research\\
	Cambridge, Massachusetts 02142\\
	Email: dgutfre@us.ibm.com
    }
    
\affil[4]{Department of Mechanical Engineering\\
	Massachusetts Institute of Technology\\
	Cambridge, MA 02139\\
    Email: faez@mit.edu
    }
    
\iclrfinalcopy

\begin{document}

\maketitle    


\begin{abstract}
In this paper, we introduce LINKS, a dataset of 100 million one degree of freedom planar linkage mechanisms and 1.1 billion coupler curves, which is more than 1000 times larger than any existing database of planar mechanisms and is not limited to specific kinds of mechanisms such as four-bars, six-bars, \etc which are typically what most databases include. LINKS is made up of various components including 100 million mechanisms, the simulation data for each mechanism, normalized paths generated by each mechanism, a curated set of paths, the code used to generate the data and simulate mechanisms, and a live web demo for interactive design of linkage mechanisms. The curated paths are provided as a measure for removing biases in the paths generated by mechanisms that enable a more even design space representation. In this paper, we discuss the details of how we can generate such a large dataset and how we can overcome major issues with such scales. To be able to generate such a large dataset we introduce a new operator to generate 1-DOF mechanism topologies, furthermore, we take many steps to speed up slow simulations of mechanisms by vectorizing our simulations and parallelizing our simulator on a large number of threads, which leads to a simulation 800 times faster than the simple simulation algorithm. This is necessary given on average, 1 out of 500 candidates that are generated are valid~(and all must be simulated to determine their validity), which means billions of simulations must be performed for the generation of this dataset. Then we demonstrate the depth of our dataset through a bi-directional chamfer distance-based shape retrieval study where we show how our dataset can be used directly to find mechanisms that can trace paths very close to desired target paths. Furthermore, we discuss how we plan to expand LINKS to include more complex mechanical components and expand the dataset in the future. our work is available at https://github.com/ahnobari/LINKS. We believe LINKS will enable a vast array of computational approaches in kinematic design.
\end{abstract}

\section*{INTRODUCTION}
The analysis and synthesis of the kinematics of different mechanisms are among the longest-standing problems in engineering design, capturing the attention of many scientists and engineers throughout history~\citep{lipson_2008}. Despite the attention on kinematic synthesis, the design of complex kinematic systems is still not well-understood and is often limited to specific tasks, which require trial and error, expert knowledge, or heuristics to find good designs. In recent decades, however, the introduction of computational design approaches has shifted the focus toward inverse kinematic design using optimization methods~\citep{lipson_2008,bacher_2015,CABRERA20021165,doi:10.1177/0954406220908616,EBRAHIMI2015189,10.1115/DETC2010-29028,ms-6-29-2015,MCGARVA1994223,10.1115/1.4001774,10.1115/1.4042325,ml1,ml2,10.1115/1.4048422,10.1115/DETC2018-85529,campbell,khan2015dimensional}. Amongst computational approaches, there is rising interest in applying data-driven approaches for inverse design. This has led to a plethora of research on the application of statistical machine learning and deep learning models in inverse kinematics~\citep{ml1,ml2,10.1115/1.4048422,khan2015dimensional} and engineering design~\citep{DBLP:journals/corr/abs-2110-10863}. Unlike optimization-based approaches which may suffer from large design space, data-driven approaches present the opportunity of learning compact representations from existing datasets. However, many computational and data-driven approaches are limited by the size of datasets available, with some only focusing on modifying existing mechanisms~\citep{bacher_2015} and others limited to specific topologies of mechanisms~(such as 4-bars, 6-bars, \etc)~\citep{CABRERA20021165,doi:10.1177/0954406220908616,EBRAHIMI2015189,10.1115/DETC2010-29028,ms-6-29-2015,MCGARVA1994223,10.1115/1.4001774,10.1115/1.4042325,ml1,ml2,10.1115/1.4048422,khan2015dimensional}. These limitations are primarily due to a very large design space --- the range of problem requirements and mechanism variations are practically limitless. As such, learning the entire design space requires large training datasets, which are hard to collect and time-consuming to simulate.


Large public datasets, such as IMAGENET~\citep{imagenet}, MNIST~\citep{mnist}, CIFAR-10~\citep{cifar}, with millions of annotated examples, have been widely attributed as one of the leading factors behind the success of machine learning approaches in computer vision. In contrast, one of the major roadblocks facing data-driven methods for the engineering design community and not just inverse kinematic design when it comes to applying deep learning approaches is a lack of large public datasets~\citep{DBLP:journals/corr/abs-2110-10863}. We note that the data used in existing data-driven kinematic design approaches is limited by both size~(most current databases include tens of thousands of mechanisms~\citep{10.1115/1.4042325,ml2,10.1115/1.4048422}) and complexity of mechanisms~(limited to 4-bars, 6-bars, \etc). 
As such, there is a need for large public datasets for kinematic design, which can enable high-performing data-driven models, provide a library of designs for practitioners to study, and establish benchmark problems for future work.

Producing very large datasets for kinematic design is a challenging problem due to the need for an appropriate representation scheme that does not waste resources in creating infeasible designs, and the large computation time required in simulating the movement of all linkages and ensuring diversity in the dataset. To address these gaps, we introduce LINKS, a dataset of 100 million one degree of freedom~(1-DOF) planar linkage mechanisms with complexity going up to 20 linkage joints. LINKS is created with a primary focus on the ``Path Generation'' problem. The path generation problem is the task of designing linkage mechanisms that generate a particular path described by a finite series of point coordinates. The dataset is made up of a large number of mechanisms and the simulated coupler paths traced by each joint of said mechanisms. However, it can be extended easily to other types of problems such as ``Function Generation'' and ``Motion Generation''~(See ~\citep{mccarthy2010geometric}). 

To produce LINKS, we overcome significant roadblocks. First, we create an efficient generation scheme that allows for randomly sampling valid mechanisms. To do this, we introduce a new operator that guarantees to create valid, non-degenerate, and non-locking mechanisms without any redundancies. We show that the proposed operator is more efficient than a widely used operator from literature~\citep{lipson_2008}. Second, we develop an efficient forward simulation algorithm, which is both vectorized and parallelized, enabling us to simulate mechanisms on a multi-core system in half a second, compared to the 454 seconds needed by a single thread non-vectorized solver. As we discuss later, the algorithm randomly samples many parameters, which leads to locking mechanisms in the majority of the simulations. For example, randomly sampling mechanisms with more than ten joints lead to a 99\% infeasible (locking) mechanism (see Fig.~\ref{fig:infeasibility}), requiring a simulator to do a hundred simulations before adding one item to the dataset. Creating an efficient simulator allowed us to generate the LINKS dataset in hours instead of requiring months. 

Another challenge faced in generating a dataset of linkages and associated coupler paths is the extreme skewness in the types of shapes obtained from all the coupler paths. We observe that two types of shapes, circles, and arcs,  make up 62\% of the paths traced. These two shapes are less interesting from the perspective of inverse kinematic synthesis~(as theoretical solutions for such shapes are easily obtained). To address this issue, we detect and filter these shapes, which leads to two datasets. One raw dataset with everything and one curated subset of paths, which randomly removes 99.5\% of these two shapes and associated mechanisms.

We believe LINKS can serve as a standard dataset for kinematic synthesis for data-driven approaches and help establish a common benchmark for future comparison of different machine learning methods. Furthermore, we hope that the introduction of such a large dataset will help garner more attention toward the kinematic synthesis problem and expedite the research and amount of information accumulated on the topic.

The key contributions of this paper are:
\begin{enumerate}
    \item We propose a new operator, named J-operator, that allows us to generate feasible mechanisms with a single degree of freedom.
    \item We release the first publicly available linkage mechanism dataset, named LINKS, with 100 million mechanisms and 1.1 billion coupler curves.
    \item We create a vectorized and parallelized forward solver, ~800 times faster than a non-vectorized solver.
    \item To show that 62\% of the coupler curves are circular and arc-shaped, revealing a large bias in the dataset. We filter the dataset to reduce this bias and identify a curated subset of 600 million coupler curves.
    \item We demonstrate a case study of mechanism retrieval by implementing a shape similarity-based search on LINKS and show that it yields accurate and diverse results, showing the efficacy of the dataset.
\end{enumerate}

\section*{ BACKGROUND \& RELATED WORK}
In this section, we provide a brief background on the path and motion generation problem in planar linkages and discuss different approaches that are typically employed in the current literature. Then, we will discuss the sequential generation of 1-DOF mechanisms and give an overview of simulation approaches in the existing literature. For a more in-depth discussion on the topic of linkages and their kinematics, interested readers are referred to~\citep{erdman1984advanced,mccarthy2010geometric}.

\subsection*{Computational Inverse Kinematics}

Computational approaches toward the inverse kinematics problem fall into three primary categories: a) Numerical atlas-based approaches, b) Optimization-based approaches, and c) Data-driven approaches. With the advent of machine learning, data-driven kinematic design is rising in popularity. While the LINKS dataset directly enables better data-driven design by providing a large and diverse training dataset, it can also help improve the first two approaches by enabling a search over a larger space or providing candidates for smart initialization of optimization methods.

\paragraph{Numerical atlas-based approaches:} The first approach is when a database of mechanisms is created and the paths produced by said mechanisms are used as a sort of a ``numerical atlas''. The atlas can be employed to look up the closest paths to any given desired path and use the associated mechanism in the existing database as a solution. This retrieval step can also be integrated with a local optimization of the mechanism to get as close to the desired path as possible~\citep{MCGARVA1994223,CHU2010867,doi:10.1080/17415977.2014.890615}. In most of these approaches, the numerical atlas is usually limited to a specific kind of mechanism or a handful of mechanism types, such as a four-bar or a six-bar mechanism. These simple mechanisms with a few joints are limited in the types of paths they can generate. In the example of four-bars, it is known that at most five points of a path can be exactly matched~(and even that is not always possible)~\citep{reuleaux1875lehrbuch}, which shows that even with a sizeable atlas, the range of possible paths that can be traced will be limited. In LINKS, we overcome this limitation by including mechanisms with a significantly higher number of joints and providing an atlas of 1.1 billion coupler paths.

\paragraph{Optimization-based approaches:}
The second approach is what we call optimization-based approaches. These include works that employ different kinds of optimization algorithms to find the most suitable mechanisms for any given target path. Although some researchers use genetic algorithms or genetic programming methods to generate mechanisms with desirable paths~\citep{lipson_2008,ms-6-29-2015}, others take the approach of using Fourier descriptors for optimization~\citep{10.1115/1.2826396,10.1115/1.4004227}. Gradient-based optimization~\citep{bacher_2015} to modify an existing mechanism is also a common approach. Apart from a few exceptions~\citep{lipson_2008,bacher_2015}, most of these approaches are either limited to optimizing existing mechanisms~\citep{bacher_2015}, or are limited to specific kinds of problems. For example, Lipson~\etal~\citep{lipson_2008}, focused on solving the straight line problem. The performance of population-based optimization approaches, such as genetic algorithms, also depends on initialization. Using the LINKS dataset, one can gain improvements in such approaches by introducing candidates that are close to the desired goal.

\paragraph{Data-driven approaches:}
More recently, machine learning-based approaches are increasing in popularity and a few data-driven works have been published on the topic. In most of these works, the previously mentioned approaches of ``numerical atlas'' and optimization are adapted to data-driven approaches. For example, Deshpande~\etal, in their paper, have adapted an approach that combined the numerical atlas approach with optimization~\citep{10.1115/1.4042325,ml2,10.1115/1.4048422}. They use variational autoencoders (VAEs)~\citep{kingma2014autoencoding} and clustering-based search to find appropriate candidates that can generate a desired coupler curve. In their other works, they employ VAEs and conditional VAEs~\citep{NIPS2015_8d55a249} to synthesize mechanisms. The datasets used in such approaches are usually small and often limited to specific types of mechanisms~(four-bar, six-bar, \etc). For instance, a dataset of 6818 linkage mechanisms is utilized in ~\citep{deshpande2021image}. Having a large dataset with millions of mechanisms can greatly benefit these models. In other data-driven approaches, researchers have approached mechanism generation by conditioning them on paths~\citep{ml1}, which is again limited to four-bar mechanisms. In contrast to these approaches which can be interpreted as the ``numerical atlas'' adaptations, other methods have attempted to implement the equivalent to the optimization approach using machine learning. One such work used deep Q learning~\citep{mnih2013playing} and Lipson's T and D operators~\citep{10.1115/DETC2018-85529}. Despite, not being limited to specific mechanisms, these reinforcement learning-based approaches need retraining for every target shape. 
Due to a lack of benchmark problems, numerical atlas-based approaches and reinforcement learning approaches have not been compared against each other for the same target coupler curves.

What is evident is that machine learning approaches show great promise, however, at the moment, two limitations can be observed in current approaches. The first is the limitation of many methods to both specific mechanisms and problem types. The other limitation is a lack of benchmark problems that can be used to compare the performance of different machine learning methods and establish the strengths and weaknesses of any new approach that is developed. The LINKS dataset addresses these limitations.

\subsection*{Sequential Generation of Planar Mechanisms And Degrees of Freedom}

\begin{figure}[h]
\vskip -0.25in
    \centering
    \includegraphics[width=\columnwidth]{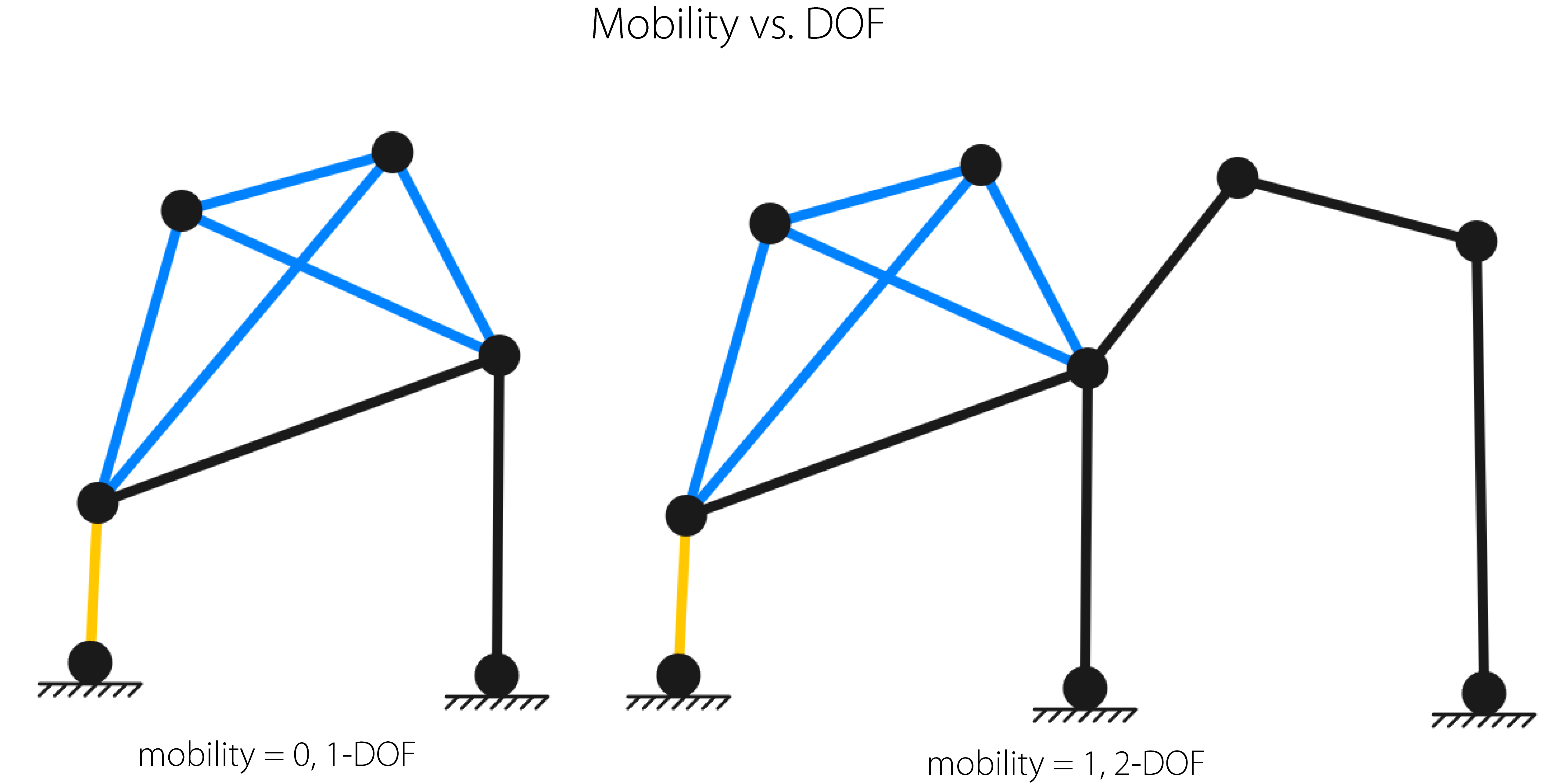}
    \caption{Two mechanisms where DOF and mobility do not match as a result of redundant linkages. In both images, one of the linkages highlighted in blue has to be removed to remove the redundancies in the mechanism.}   
    \label{fig:dofmobility}
\end{figure}

LINKS consists of planar linkage mechanisms with 1-DOF. It is crucial to discuss the existing approaches in generating such mechanisms before we introduce our approach. First, we must establish the difference between mechanisms with one degree of freedom and mechanisms with mobility of one. Unlike degrees of freedom, which can be ambiguous, the mobility of a planar linkage mechanism is clearly defined and easily measurable~(See~\citep{erdman1984advanced,mccarthy2010geometric} for more details). The equation of mobility is as follows:
\begin{equation}
    m=3(n-1)-2j_1-j_2
\end{equation}

where $n$ is the number of linkages and $j_1$ is the number of 1-DOF pairs (such as joints or hinges in planar mechanisms) and $j_2$ is the number of 2-DOF pairs~(such as cams, wheels, \etc). Since we focus only on planar linkage mechanisms with simple joints/hinges, we can ignore the $j_2$ term in this equation. A mechanism with mobility of zero or negative mobility can be 1-DOF~(an example is shown in Fig.~\ref{fig:dofmobility}), and there can be mechanisms with mobility of one but degrees of freedom larger than one~(see Fig.~\ref{fig:dofmobility}), these happen when redundant linkages are added to mechanisms. Since the focus of our dataset is on ``path generation'', we build our dataset such that there are no redundancies, meaning that mechanisms in our dataset are guaranteed to be 1-DOF and have the mobility of one at the same time.

An approach to generate 1-DOF mechanisms was proposed by Lipson~\citep{lipson_2008} and later also used by Vermeer~\etal~\citep{10.1115/DETC2018-85529}. Lipson~\citep{lipson_2008} proposed two operators for editing 1-DOF mechanisms such that the mechanisms would retain 1-DOF. He called these operators the T and D operators which operated on linkages and proved that these operators will retain the original DOF of the mechanism~\citep{lipson_2008}. The D operator simply added a joint with two linkages connected to the ends of the linkage being operated on and the T operator broke down the linkage being operated in two. A limitation of their approach is that the algorithm will have to start from an existing 1-DOF mechanism and the possible mechanisms that can be generated in such a method are, therefore, sensitive to the initial mechanism. We propose an approach to operate on entire mechanisms based on adding joints which describe in the methodology section.


\subsection*{Simulation of Kinematics}
There has been substantial work done in solving 1-DOF mechanisms, however, as mechanisms get more complex, solving them becomes costly and the complexity of the closed form analytical equations becomes gargantuan. As a result, algorithms-based and numerical approaches to solving such systems are typically employed~\citep{10.1115/1.2826898,10.1115/1.4046817}. The literature on this topic is extensive and beyond the scope of this paper, however, there are a few relevant works that we will discuss here as they set up the context for future discussion. Broadly, two different approaches can be considered in simulating mechanisms beyond analytical approaches~\citep{10.1115/1.4046817}. One approach is the numerical approach to solving kinematic systems. An example of such an approach is Lipson's simulator, used for the genetic programming approach in ~\citep{10.1115/1.2198255}. In most numerical approaches planar mechanisms are solved using numerical algorithms used for solving systems of non-linear equations~(such as Newton-Raphson or Broyden's method), these approaches are capable of simulating very complex systems, however, in many complex systems the solution is not unique and these simulators only produce one of the possible results~\citep{10.1115/1.2826898,10.1115/1.4046817}. The other approach to solving linkage mechanisms is to take a graphical approach and solve planar mechanisms from a purely kinematic approach. One such simulation approach which focuses purely on kinematics is the one proposed by B\"{a}cher~\etal~\citep{bacher_2015}. The simulator proposed by B\"{a}cher~\etal solves a linkage system iteratively by starting from known values such as the position of the ground joints and the current position of the actuator arm and solving for any joints that can be solved with the available information~(taking into account initial positions). At every iteration, more joints will be solved until at the final iteration where all joints are solved. This approach is illustrated in Fig.~\ref{fig:solver}. While these approaches are limited to mechanisms with simple kinematic loops consisting of dyadic loops, an advantage of using them is that the gradients of simulation can be obtained in a similar manner which enables gradient-based optimization~(\eg, editing existing mechanisms to fit certain constraints~\citep{bacher_2015}). We adopt the approach of B\"{a}cher~\etal and discuss the details of our work in the following sections. For a more in-depth view of simulation methods, readers are referred to~\citep{10.1115/1.4046817}.


\section*{ LINKS: DATASET \& METHODOLOGY}
\begin{figure*}[h]
    \centering
    \includegraphics[width=\columnwidth]{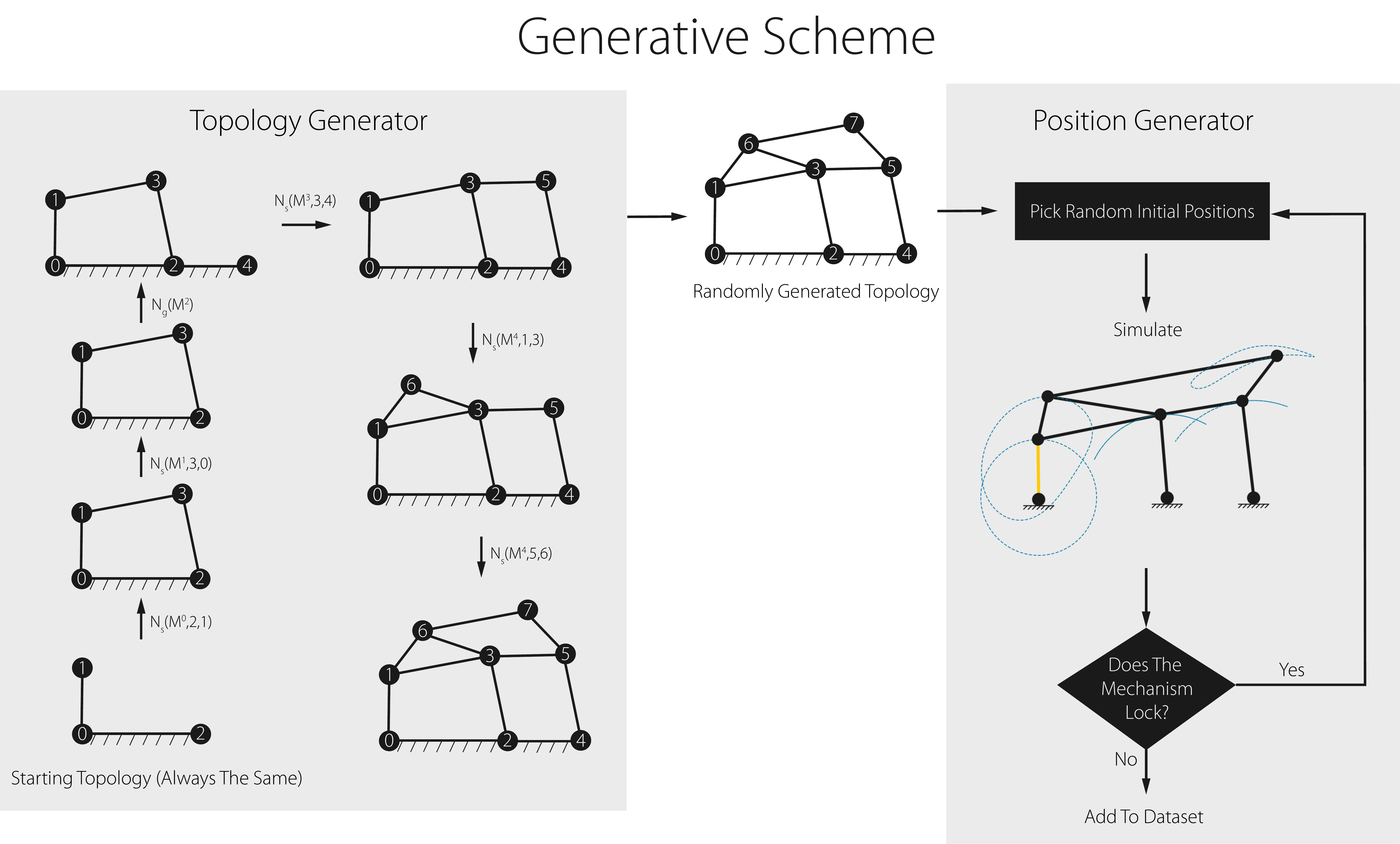}
    \caption{Overview of the process used to generate planar linkage mechanisms. Note that the initial mechanism used in the topology generator is the simplest 1-DOF mechanism possible. It is important to mention that for each topology, the position generator is run multiple times to obtain multiple candidates.}   
    \label{fig:overview}
\end{figure*}

We create a dataset of 100 million planar linkage mechanisms with 8-20 nodes resulting in over 1 billion different coupler curves.
This scale of data generation comes with notable challenges. We implement hardware and software engineering improvements, to create this dataset within a practical amount of time. In this section, we will describe our dataset features as well as the methodology used to generate it.

\subsection*{1-DOF Operator For Planar Linkage Mechanisms}
To generate mechanisms with mobility of one there are several approaches. The naive approach would be to randomly generate graph topologies~(\ie, randomly connect a number of joints without mobility in mind) and measure their mobility and repeat the process until appropriate topologies are identified. This naive approach is inefficient as the number of all possible topologies is extremely large for large enough mechanisms. Therefore, an efficient approach must be established to enable the fast generation of valid mechanisms.

\paragraph{J operator:} We propose a simplified and unified single operator to edit existing mechanisms, named, joint, or J operator. In our approach, rather than focusing on the linkages for operations~(in contrast to T and D operators~\citep{lipson_2008}), we apply operations on two existing joints instead. The simple operator that is used to generate new mechanism topologies simply picks two existing joints within any given mechanism and adds a new joint that has linkages connecting it to the two joints that were selected~(See Fig.~\ref{fig:overview}). Assuming the initial mechanism has mobility of 1~(\ie, $m=1$), adding two linkages to the system means adding 2 to $n$ in the mobility equation, however, the operation also adds 3 $j_1$ pairs~(\ie, the added joint and the two other joints which just received two new connections) which essentially retains the mobility of the initial mechanism. Furthermore, since mobility is maintained in this manner, so long as the initial mechanism has no redundant linkages and constraints, the resulting mechanisms will also have no redundancies either~(since mobility is maintained). We will denote an operation as $N(M^{t},i,j)$, where $M^t$ is the current mechanism at iteration $t$ starting at $t=0$ with the stated initial configuration and $i,j\leq t+3$ referring to the index of the nodes in the mechanism, which the operator is applied to. Similar such approaches have been explored and theorized by many before us and are considered a robust method of building mechanisms and performing inverse kinematic design~\citep{suh_radcliffe_1987}.

\paragraph{Ensuring simple kinematic loops:} Our operator also ensures that the resultant mechanism always has simple kinematic loops. Our operator is a dyadic operator, where every operation of our operator creates a simple kinematic loop between the new joint and the two existing joints. Therefore, as long as the initial mechanism consists only of simple kinematic loops, any resulting mechanism will also have only simple kinematic loops. This is useful for our simulations, as we use graphical solvers which require simple kinematic loops.
The proposed operator could be used in systems with more degrees of freedom as well and would retain the mobility of the existing mechanism. Additionally, this approach could easily be extended to 3D linkage systems~(with universal joints) as well with the operator now operating on three joints to maintain the original mobility. 



\paragraph{Initialization:} Finally, it is important to discuss the initialization of the sequential generator. The mechanism used at the start of the sequential generator will be the simplest possible 1-DOF mechanism---a single actuator comprising of a ground joint and a linkage, which is the actuator arm~(See Fig.~\ref{fig:overview}). Besides the actuator arm, our initialization mechanism also consists of a floating ground joint which is fixed in space~(See Fig.~\ref{fig:overview}). We start from this point as this is the simplest possible 1-DOF mechanism and therefore any generated mechanism from this point on will not be biased by an initial mechanism as no simpler mechanism is possible and future mechanisms will only be generated based on the order by which the operator is applied.

\subsection*{Graphical Simulator}

\begin{figure}[h]
    \centering
    \includegraphics[width=0.6\columnwidth]{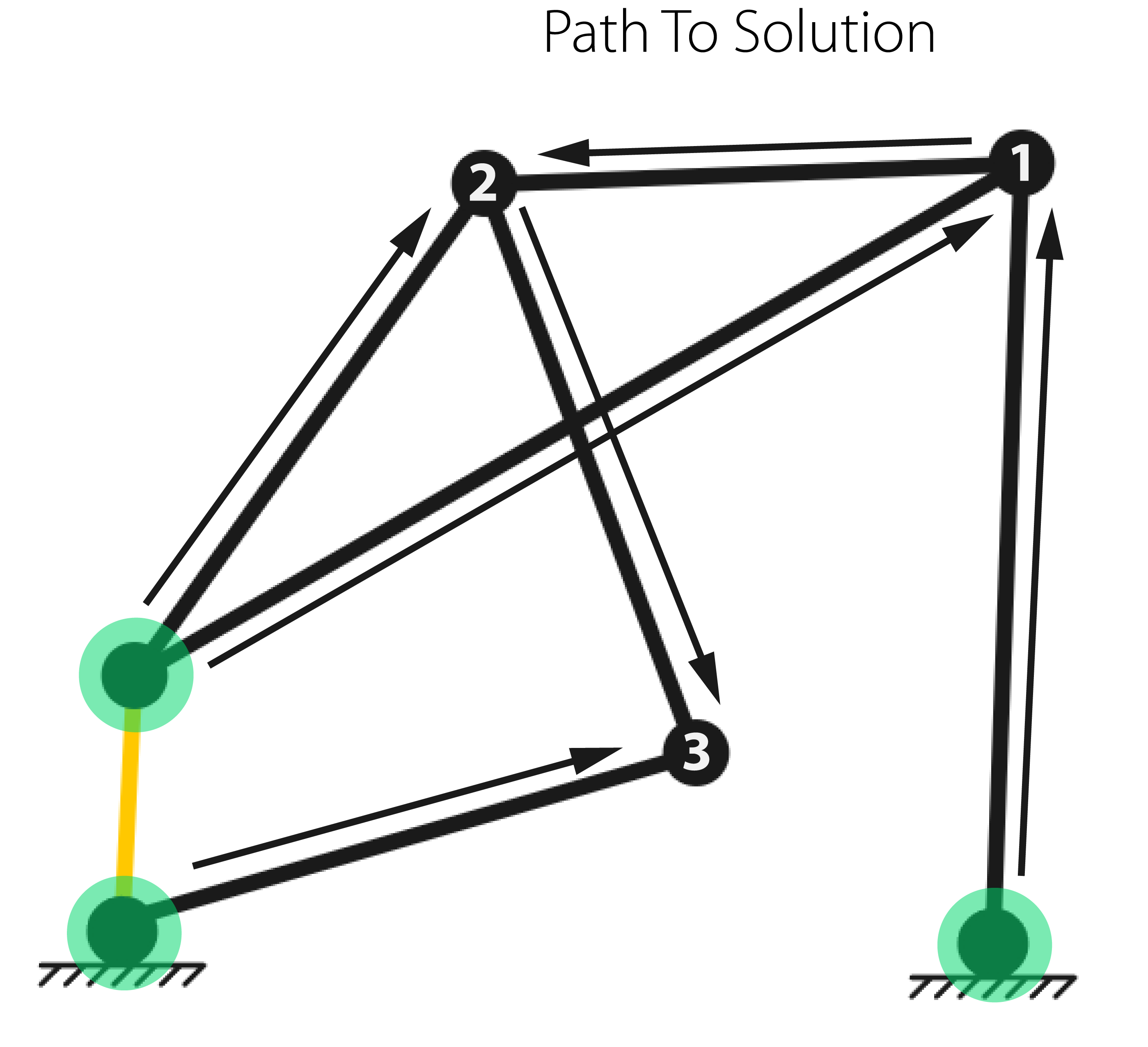}
    \caption{Here we illustrate the path the solver takes to find the solution. At first, the solver starts with the known joints~(\ie, fixed and actuated joints) and at every step nodes with two known neighbors can be found, in this mechanism illustrated, the path to the solution has 3 steps. The numbers of the joints indicate the order in which the solution is found and the arrows indicate which two neighboring joints are needed to solve the given joint. Known nodes are highlighted in green.}   
    \label{fig:solver}
\end{figure}

We adopt the solver proposed by B\"{a}cher~\etal~\citep{bacher_2015}. 
In the approach, the authors take any given mechanism with simple kinematic loops and rather than performing a dyadic decomposition~\citep{10.1115/1.4046817} to identify four-bar loops, take an iterative approach to find the solutions by modeling mechanisms as graphs. Starting from joints with known positions~(actuator arm and fixed joints), they solve for any joint which has two known neighbors~(\ie, two joints with currently known positions that have linkages connecting to the node we are trying to solve). They then repeat this iterative scheme until all joints have known positions. In doing this a path to the solution of all joints is found which can be used to find solutions at different positions of the actuator. Note that the solver will fail if the mechanism is made up of complex kinematic loops. 

When generating a large dataset, once a topology is generated using our dyadic operator, many candidates of initial positions have to be evaluated to find joint positions which do not lead to locking mechanisms. Furthermore, for each mechanism topology, many candidates are needed to capture the output space for any given topology. This means that the same mechanism topology will have to be simulated thousands of times. Re-initializing the solver every time and finding the path to the solution is not efficient. We overcome this issue, by using the following approach. Once the path to the solution is known, we employ this path to solve different variations of the same topology without any more neighborhood searches.  This significantly improves the speed of simulations compared to performing searches for every candidate. 

The solution paths also allow effective gradient-based optimization, which we do not discuss in this paper for brevity.
For details on how the path to the solution is identified see Fig.~\ref{fig:solver}. We have created an interactive website based on the aforementioned solvers (\url{http://decode.mit.edu/links}) which allows users to generate their own mechanism topologies, edit mechanisms, and simulate the results in real-time. Our hope is that practitioners with little computational experience can also use the website to design and simulate new mechanisms.

\subsection*{ Mechanism Representation}

In our approach, we model mechanism topologies as undirected graphs, in which each joint~(regardless of type) is considered a ``node'' and each linkage is considered an ``edge''. This representation is inspired by the solver, where graph-based walks can be used to solve planar linkage mechanisms exactly. To complete the definition of the graphs, each of the nodes will have features that include their type~(fixed, simple, actuated) and initial positions. Each graph is represented by an $n \times n$  adjacency matrix $A$, who's $(i,j)$ entry is zero if nodes $i$ and $j$ are not connected and one if they are connected. Finally, each mechanism is also represented by a feature matrix which in our representation is an $n\times3$ matrix. Each row of this matrix includes the features of each node. The first element of each row indicates the type of the node~(\ie, 1 for simple node, 0 for fixed node, and 2 for actuated nodes), and the two other elements in each row are set to be the initial positions of the nodes at time zero~(which can be used to determine the length of linkages between the joints). We believe graph-based representation can provide a unified approach for representing all kinds of kinematic systems with different types of components and relations.


\subsection*{Dataset Generation Approach}
To generate a very large mechanism of planar linkages, we take a random sampling approach. This is analogous to stochastically searching within the design space. The J operator enables us to create mechanisms consisting only of simple kinematic loops. 
Besides the topology, initial positions are also needed to simulate a mechanism as different initial positions can lead to completely different shapes of coupler paths. Therefore, we separate our generation process into two distinct steps. The first step is to generate valid 1-DOF topologies. The next step is to come up with different initial positions for each of the topologies such that the resulting mechanisms are not locking (valid). An overview of our approach is illustrated in Fig.~\ref{fig:overview}.

\subsubsection*{Generating Valid Topologies}
In order to create valid topologies, we make a slight modification to our single operator. Note that if the operation is applied to two ground nodes~(\ie, add a node connected to two existing ground joints) the generated joint will be effectively a fixed ground joint. In practice, however, this reduces the probability of ground joints appearing. This is because only two nodes at the start will be ground nodes, and, with large mechanisms, the odds of randomly selecting both nodes~(which is what is needed to add a new ground node) will be slim. As a result, we modified our operator into two separate entities which allow us to control the probability associated with nodes being fixed. To do this we break down, the original operator $N(M^t,i,j)$ into $N_s(M^t,i,j), \{i,j\}\not\subset Fixed Joints$ and $N_g(M^t)$, where $N_g$ simply adds a fixed node and $N_s$ does not operate on two fixed joints simultaneously. During the topology generation process, we simply associate probabilities to each of the two operators. In our dataset generation process, we set the probabilities such that the likelihood of $N_g$ is one-third of the probability of $N_s$~(See Fig.~\ref{fig:overview}).

Now that the basic scheme of mechanism generation is described, we list the basic rules utilized during mechanism topology creation:

\begin{itemize}
    \item[$\bullet$] The number of nodes in any mechanism topology will be sampled from a uniform distribution ranging from 8 to 20 nodes. $n \sim U(6,15)$.
    \item[$\bullet$] The assumption will be made that the actuator arm will always be the linkage present in the initial configuration before any operations are applied.
    \item[$\bullet$] Once the number of joints, $n$, is selected, the generated mechanism shall not be reducible to a mechanism with fewer joints. Thus, the kinematics of the mechanisms topologies that are generated will rely on solving the positions of all nodes, and the final joint~(in the solution path) will not be determined without all other joints having been solved~(\ie, the solution path of the graphical solver to the final joint must go through all joints). The reason this rule is imposed for topology generation is that this allows for uniform coverage of the design space for mechanisms with 8-20 nodes, this is due to the fact that, if a mechanism with 15 nodes was separable into two separate mechanisms of smaller size, the mechanism in question would be practically representative of the mechanisms with smaller size rather than representing the space of mechanisms with more complexity and larger size.
    \item[$\bullet$] The final joint in the solution path shall not be connected to any fixed joint. This rule is imposed to prevent complex mechanisms with high numbers of joints from producing simple paths such as arcs~(which is what a linkage connected to one fixed joint will always result in). Because the focus of our work is on generating mechanisms with path synthesis in mind, having a complex mechanism that ends up producing arcs is not desirable. Furthermore, mechanisms with many nodes produce many shapes~(one trace per node is simulated) and many of the nodes within a larger and more complex mechanism are already producing arcs to begin with, which means that this rule does not come with the loss of generality.
\end{itemize}

To ensure all of the above rules are met, besides the operator effectively maintaining mobility of one and avoiding redundancies, we employ some simple heuristics when applying operators during the generation process. 

\subsubsection*{Finding Valid Initial Positions}
Once different topologies are generated we must find appropriate initial positions for the joints to complete the definition of mechanisms and simulate each mechanism to obtain the paths generated by each mechanism. For our dataset, we set some basic rules for what is a valid mechanism with respect to initial positions:

\begin{itemize}
    \item[$\bullet$] Bounding box: All initial positions of mechanisms must fit within 1x1 space. During the random position selection, positions are sampled from a uniform distribution ranging from 0 to 1 in both the $x$ and $y$ directions. $x_0 \sim U(0,1), y_0 \sim U(0,1)$.
    \item[$\bullet$] Infeasibility: A mechanism is considered valid if the actuator arm can make a full rotation without the mechanism locking. Although in many cases researchers have focused on the path traces, even when the full rotation of an actuation arm was not possible and looked at rocking actuators or specific functions being imposed by actuators, we take the most generalizable path with the least assumptions to generate mechanisms which allow for a full rotation of the actuator arm without restriction.
    \item[$\bullet$] Translation invariance: The actuator arm is always placed at the center of the 1x1 domain meaning the first ground joint is always located at $(0.5,0.5)$. We impose this rule as a form of standardization with respect to translation in space. This is to say that by doing this we ensure the same mechanisms translated in space cannot exist within our dataset.
    \item[$\bullet$] Scale invariance: The length of the actuator arm is set to be a fixed length of 0.05. Similar to the previous rule, this also imposes a standardization with respect to scale. By setting the actuator arm to be of the same length, we impose scale in-variance within the data.
    \item[$\bullet$] Rotation invariance: The actuator arm is always set to be horizontal at the initial position which standardizes the dataset with respect to rotation. In turn, making the dataset rotation invariant, and an actuator invariant~(Note that the same mechanism at a different actuator position can have different positions which can be taken as initial positions without changing the underlying kinematics, and by setting the actuator to always be horizontal the dataset will be invariant to this matter).
    \item[$\bullet$] Position Variety: Each topology must be represented with five different initial positions. This means there will be five variants of any mechanism topology within our dataset. This is valuable as this will provide a more complete picture of the output space of any given mechanism topology, compared to only having one valid configuration for each topology.
\end{itemize}

Similar to the case of topology generation, these rules are imposed using some simple heuristic algorithms. It is important to note, however, that the primary difficulty with initial position selection in comparison to topology generation comes with the second rule. Once random positions are determined with respect to the other rules, these positions must be evaluated by simulating the mechanism through an entire rotation of the actuator and if a mechanism is locking, the process will have to be repeated and the simulation time is wasted. Despite the improvements in the simulation discussed before, this remains the most time-consuming part of the task and the primary bottleneck in generating large datasets. In the following section, we will discuss how we cope with this and other challenges.

\subsection*{Challenges In Large Kinematic Dataset Generation}
In this section, we will briefly describe the challenges in generating a large dataset and our solutions to address these challenges.

\subsubsection*{Scale of Infeasibilities}
\begin{figure}[h]

    \centering
    \includegraphics[width=0.8\columnwidth]{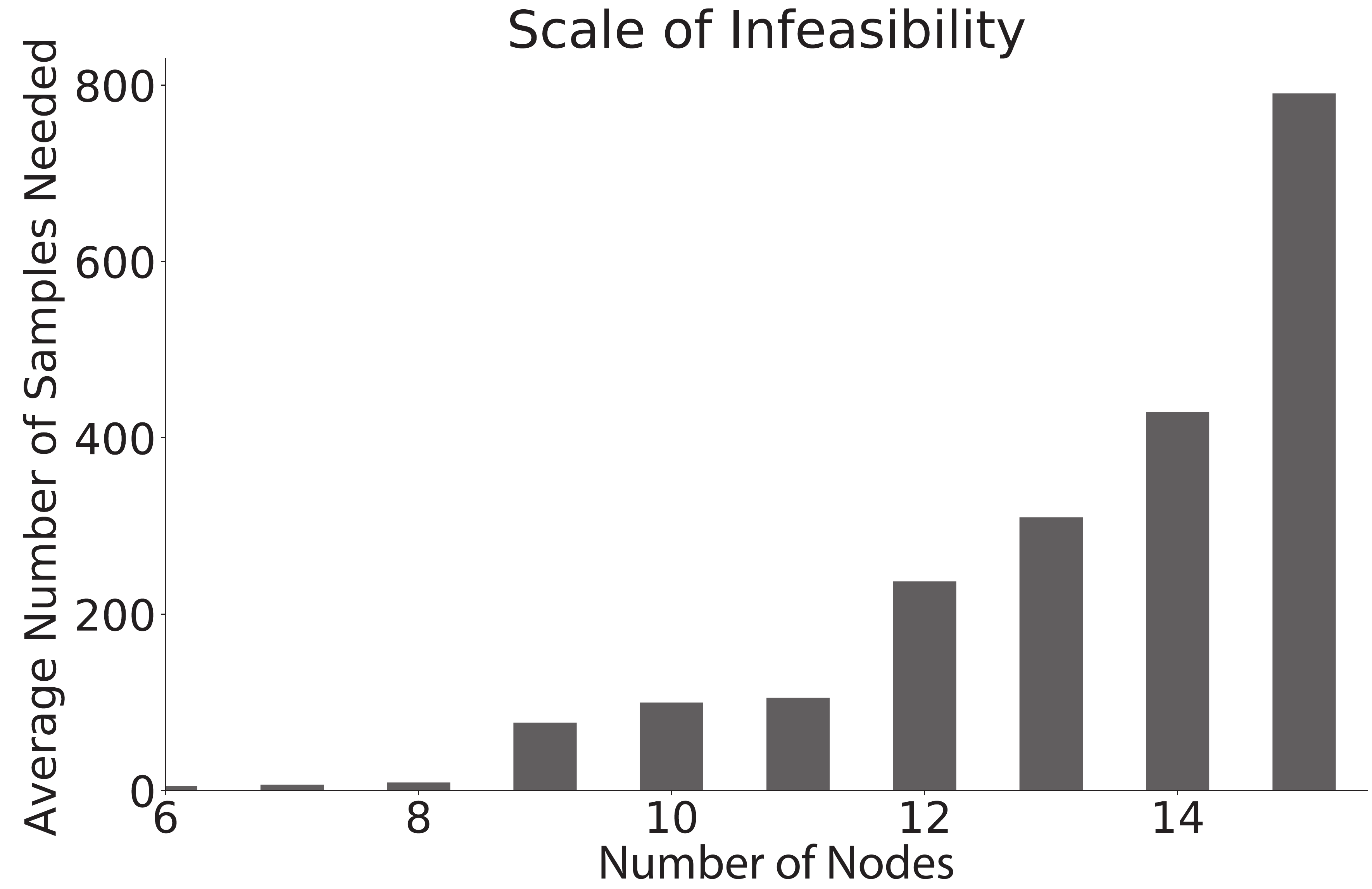}
    \caption{Average number of random samples needed to find a valid solution for mechanisms of different sizes. The figure shows that as the number of joints increases, it becomes increasingly difficult to find a valid mechanism, with $\ge 99\%$ samples wasted for mechanisms with more than 10 nodes. On average, more than 500 simulations are needed for each feasible design.
    }   
    \label{fig:infeasibility}
\end{figure}
As discussed before, the scale of our dataset is significantly larger than any existing public dataset, thus the number of mechanisms that must be generated for such a dataset is extremely vast. This would only be possible if finding new mechanisms was a speedy enough task to make generating such a large dataset of mechanisms possible. The primary issue in generating new mechanisms randomly comes from the scale of infeasibility. As mentioned before, the first difficulty comes from finding mechanism topologies that are 1-DOF and do not include any complex loops. The number of different topologies~(assuming order matters) with ``n'' being sample nodes and ``m'' being ground nodes, with one actuator node is in the order of $O(2^{n(n+m)})$. Given that most of these topologies would not be valid for our purposes, the task would become quickly impractical. We overcome this by using our topology generation based on a single dyadic operator which facilitates the generation of topologies that are guaranteed to meet our criterion~(Note that it can be proven that the dyadic operator is capable of producing all possible 1-DOF mechanisms without complex kinematic rules). In this way, the first aspect of infeasibility has been overcome.

We sample initial positions from a uniform distribution, meaning the space of possible values for positions is continuous, therefore, there are infinite possibilities for positions. What is more concerning is the fact that there are no simple approaches such as the dyadic operator which can overcome this issue easily, therefore we have no choice but to sample randomly every time until valid positions are found. Furthermore, the more complex the mechanisms get~(\ie, the more joints there will be) the harder it becomes to not produce locking mechanisms, as only one slight change in position within the system can result in a locking mechanism. We show this difficulty empirically by taking random topologies of different sizes and measuring how many random samples are needed before a good initial position is found. As expected, on average, the larger the mechanism, the more difficult the task is, and the more random sampling required before a solution is found~(See Fig.~\ref{fig:infeasibility}). Additionally, we found that the infeasibility increases with size in a non-linear and exponential way. 

\paragraph{Low fidelity evaluations for infeasibility detection:} In our dataset generation, we found that on average, for the entire dataset, we required evaluation of more than \textit{500 random positions} for each candidate. This means for a dataset of this scale the number of evaluations quickly becomes infeasible. To be able to produce such a dataset, we must optimize the simulation, and the random generation process in some way. We discuss details on scaling simulations and dealing with the high computational costs later. However, we do implement some basic heuristics to improve the efficiency of random sampling. To ensure mechanisms are not locking over the entire revolution of the actuator arm we must simulate the mechanism over a large number of actuator positions, however, this is costly as many simulations must be performed, therefore, when sampling randomly we only evaluate the mechanism on 50 separate equally spaced positions of the actuator rather than the 200 we use for a more smooth simulation of the mechanisms. Once valid positions are found for the 50-point simulation, we perform the more refined simulation of 200 points to verify that the mechanism is not locking, and if the mechanism fails in the high fidelity simulation we continue random sampling. The logic behind this move is that most mechanisms that do not lock in the low fidelity simulation are likely to be valid in the high fidelity simulation as well. Therefore, by doing this, we reduce the cost of each evaluation by four times and significantly improve the time required to find valid positions for a given topology.

\subsubsection*{ Computational Cost of Simulations}
For the generation of a dataset of this size, we must scale and improve our simulation approach. To do this, we take two primary steps. The first step comes in the simulation algorithm itself. Since we are simulating the same topology many times, we set up our simulator such that the path to the solution is obtained only once and the path of simulation is used to make a vectorized version of the simulations function~(Note that the simulation algorithm itself is not vectorizable in general but for a specific topology the path to the solution is known and therefore a vectorized solver specific to a given topology can be found). Obtaining a vectorized version of the function comes at a negligible computational cost as the same function is used many times for many different actuator positions and different initial positions.

Despite these improvements, the process of performing this many simulations is still slow, and generating such a large dataset can require weeks. To overcome these limitations and make the process of generating a dataset of such size practical, we parallelize the entire process of generating mechanisms. Using the supercomputing resources provided by MIT SuperCloud and Lincoln Laboratory Supercomputing Center, we are able to run up to 640 threads simultaneously on 8 computing nodes each providing 80 threads. By scaling more than 100 times, what could be achieved in a week, can be achieved in hours and what requires months becomes a matter of days. To demonstrate the effectiveness of this approach, we provide experimental data for the time required to simulate 100,000~(10 experiments for 10,000 mechanisms) random mechanisms with single thread non-vectorized solver, single thread vectorized solver, and the parallelized and vectorized approach (on 80 threads, for consistency in experiments only one computing node was utilized for this experiment). The results of the average time over 10 experiments are provided in Table.~\ref{table:timing}. As shown, parallelizing the process hugely increases the efficiency of generating a very large dataset of mechanisms. This means that we are able to produce our dataset and expand it in the future with great efficiency.

\begin{table}[h!]
\centering
\caption{Experimental results on time taken to simulate 10,000 mechanisms averaged over 10 experiments. Values after $\pm$ indicate standard deviation.}
\vskip 0.1in
\label{table:timing}
\begin{tabular}{cc}
    \hline
    Method & Time Elapsed(s)\\
    \hline
    Single Thread Non-Vectorized Solver & 454.864 $\pm$ 7.071\\
    Single Thread Vectorized Solver & 10.994 $\pm$ 0.2910\\
    Parallelized Approach(80 Threads) & 0.5557 $\pm$ 0.0729\\
    Parallelized On A Single V100 GPU & 0.1823 $\pm$ 0.0261\\
    \hline
\end{tabular}
\vskip -0.1in
\end{table}

\subsection*{Dataset Curation \& Processing}

\begin{figure}[h]
    \centering
    \includegraphics[width=\columnwidth]{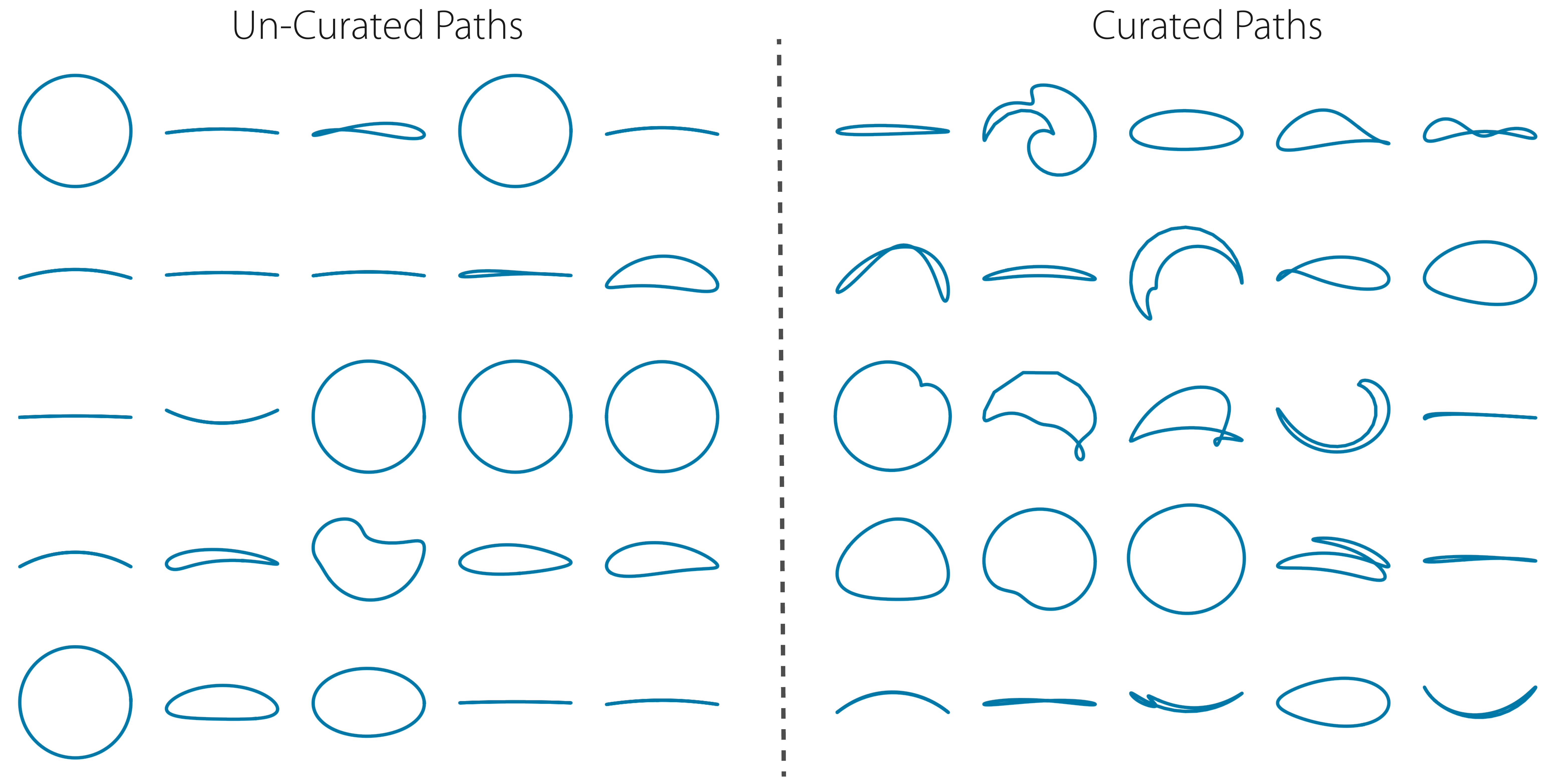}
    \caption{Random subsets of normalized shape before~(left) and after~(right) reducing arcs and circles. Without curation, the majority of paths are circles and arcs on the left side, which introduces a huge bias in the data towards such shapes.}   
    \label{fig:curation}
\end{figure}

Once the dataset is produced and all of the mechanisms are simulated, the total number of curves produced by our dataset is over a billion paths. Besides providing the raw simulation data with each mechanism, we provide a dataset of paths produced by all mechanisms with the index of their corresponding mechanism and the index of the joint within the mechanism in question. Furthermore, we provide a simplified algorithm that can reduce any mechanism down to only the joints needed to produce the needed motion for any of the curves within the dataset. To allow for easier processing and analysis of these paths, we come up with a normalization scheme for paths as well to standardize the paths dataset. In our approach, we remove scale, orientation, as well as the translation in paths. This is because, if all paths are standardized in this way, they can be compared irrespective of their location in space, scale, and orientation. To do this, we provide a normalization algorithm that takes a path with finite points, finds the line of maximum distance on the path, orients the path such that this line is horizontal, then the algorithm scales the path such that the maximum length~(which was just found) is equal to 1.0 and positions the resulting path in the center of a 1x1 box. By doing this all shapes are standardized~(examples of normalized shapes can be seen in Fig.\ref{fig:ret}).

Finally, it is important to discuss the imbalances in the paths generated by random planar mechanisms. The paths created by all the mechanisms tend to be biased heavily towards two specific kinds of shapes: arcs and circles. The reason these shapes are very common is that any node connected to the ground produces arcs. Circles are also very common since the nodes connected to the rotating arms of the actuator produce circles. Since having a path dataset made up of large amounts of circles and arcs would be inefficient for path generation problems and can introduce heavy bias which can weaken data-driven approaches, we curate our dataset of paths to remove most arcs and circles. In fact, we remove 99.5\% of arcs and circles and only keep 0.5\% of them in the curated dataset. In order to detect arcs, we simply look at the mechanism and see if the joints producing the path are connected to the ground or not. To detect circles, we look at the normalized path and measure the variance of the distances of all points in a path from the center of the 1x1 box~(A perfect circle will have no variance). We consider paths with a variance smaller than $5\times10^-4$ to be close enough to a circle and remove such candidates in the curated dataset. We find that 62\% of the paths if not curated would fall under the category of circles and arcs~(circles make up 34\% and arcs 28\%), which shows the importance of this step. To demonstrate this Fig.~\ref{fig:curation}, shows a few randomly selected normalized paths from both the un-curated as well as the curated dataset.

\subsection*{Dataset Features \& Statistics}


Here we will go over some of the features and statistics of our dataset that will be released publicly upon the acceptance of this paper. Our dataset has seven components listed below:
\begin{enumerate}
    \item\textbf{Mechanisms}: The dataset includes 100 million different mechanisms. These mechanisms are described by the adjacency matrix of the graph describing the topology of the mechanism, and a $n\times3$ matrix describing the initial positions and types of the mechanism joints. The assumption in these mechanisms is that the actuated arm is the arm connecting joints 0~(always fixed) and 1~(always simple actuated joint). The size of these mechanisms in the dataset is between 8 joints and 20 joints.
    \item\textbf{Simulation Data}: For each of the 100 million mechanisms in the dataset, we simulate the kinematics of the mechanisms for 200 equally spaced positions over the full rotation of the actuator arm.
    \item\textbf{Raw Normalized Paths}: We normalize all the produced paths and store the index of mechanism and joints associated with each path.
    \item\textbf{Curated Normalized Paths}: The same as the normalized dataset but with 99.5\% of circles and arcs removed.
    \item\textbf{Negative Samples}: 100 million mechanisms but in locking configuration~(\ie, initial positions). These mechanisms may be useful for studies on understanding locking mechanisms.
    \item\textbf{Code}: The code to create the dataset, simulate mechanisms, reduction of mechanisms down to specific joint kinematics, and path normalization will be provided. 
    \item \textbf{Web demo}: An interactive web-based user-interface is also developed and accessible at \url{http://decode.mit.edu/links}.
\end{enumerate}

The dataset and code will be made public upon the acceptance of this work at \url{https://github.com/ahnobari/LINKS}. 

\subsection*{Potential Use-Cases of LINKS}
It is critical to discuss how such a large dataset can be useful for both the design and kinematic synthesis community. As discussed in the background section, many approaches in inverse kinematics of planar linkages revolve around the numerical atlas and optimization approach. When it comes to numerical atlas methods~\citep{CHU2010867,doi:10.1080/17415977.2014.890615,10.1115/1.4001774,MCGARVA1994223} and data-driven adaptations analogous to numerical atlas methods~\citep{10.1115/1.4048422,10.1115/1.4042325,ml2}, given the fact that existing numerical libraries are limited and include mechanisms in the order of 10s of thousands ~\citep{10.1115/1.4048422,10.1115/1.4042325,ml2,CHU2010867,doi:10.1080/17415977.2014.890615,10.1115/1.4001774,MCGARVA1994223} the advantage of such a large dataset is can be easily recognizable. Moreover, when it comes to optimization-based approaches the mechanisms and shapes within our dataset can be useful for initial guesses, or in the case of genetic programming and genetic algorithms, they can be beneficial as a database for initialization and sampling. Thus allowing such algorithms to start from a valid and diverse set of mechanisms rather than randomly initialized populations that may be invalid from the beginning. At last, perhaps the most glaring benefit of our work is going to be the contribution of such a large dataset to data-driven approaches which have been demonstrated to scale well with big data. A famous example of massive data, which lead to a huge leap in state of the art is the GPT models developed by OpenAI~\citep{brown2020language}. In data-driven approaches, our dataset can be utilized in two ways which encompass both forward and inverse design modeling. Thus far, we have focused most of our discussion around inverse kinematic design but as aforementioned, the simulation process is an important part of kinematic analysis and synthesis. Although solvers like ours are differentiable~\citep{bacher_2015}, the gradient obtained by such solvers is only stable in the neighborhood of a valid mechanism and has significant discontinuities~(\eg, when locking mechanisms are present and the solution and gradient, do not exist), in turn making the direct use of these simulators in gradient-based optimization and machine learning training impractical. Therefore, a relaxed but accurate data-driven forward model for kinematics of planar mechanisms can be extremely valuable for optimization and data-driven approaches.

\section*{A CASE STUDY OF MECHANISM RETRIEVAL}
\begin{figure*}[htp]
    \centering
    \includegraphics[width=\columnwidth]{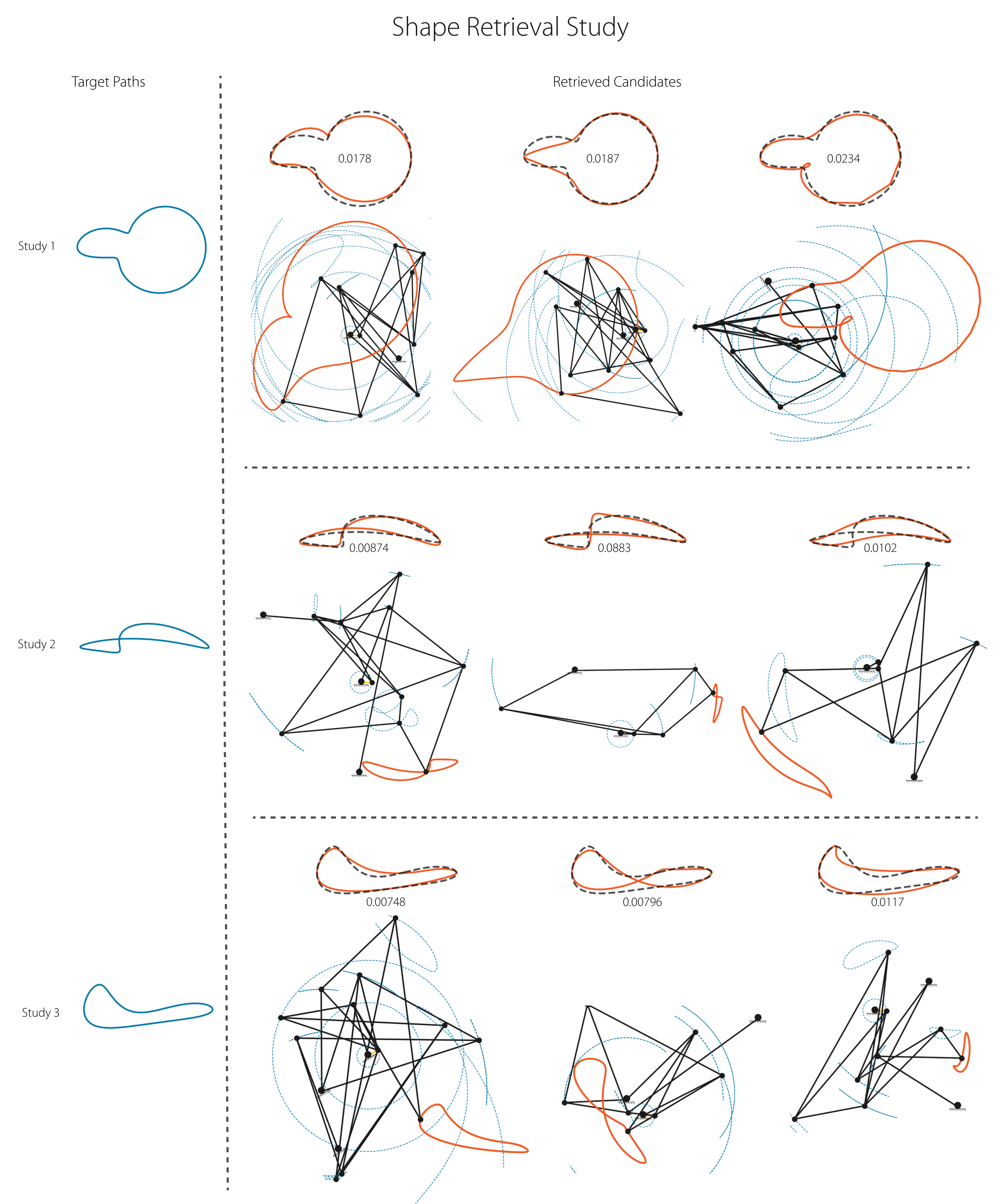}
    \caption{Results of three path retrieval studies performed on our dataset. Chamfer distance between the target and retrieved paths are displayed below/on the comparison of normalized curves.}   
    \label{fig:ret}
\end{figure*}

One of the straightforward uses of LINKS is its application in creating a numerical atlas or a library of candidates for mechanism retrieval. However, as our dataset contains more than a billion shapes, comparing any target path to all shapes would be an arduous and impractical approach. In this section, we present a simplified baseline approach to mechanism retrieval. 

Our goal is to demonstrate the depth of our dataset's coverage by selecting random shape queries and identifying mechanisms in the dataset that can create a shape that is similar to the query shape.
We first come up with challenging problems, by generating new random mechanisms~(as we did for the dataset) and picking a sample of paths from them as queries. Note that these paths do not exist in the dataset. We use the chamfer distance to compare any two normalized shapes to our targets and set a threshold below which we consider a path close enough to a given target query~(chamfer distance below 0.03~\footnote{This cut-off was chosen based on visual inspection of similarity between two curves. A lower value of the cut-off will make it less likely to find a shape that is similar to the query shape}). Our strategy for shape retrieval is to shuffle the curated path dataset and go through the data until three candidates are found for any target shape. 

For this study, we choose three query paths and run a search on a reduced dataset of ten million curated paths based on bi-directional chamfer distance (a metric commonly used to compare two point clouds in computational geometry). The top 3 best matching results for each shape are then presented in Fig.\ref{fig:ret}. As evident, even without any optimization, a simple search yields mechanisms that meet the target and are different from each other. 
This simple case study demonstrates the depth of our dataset and clearly demonstrates that the dataset, if used properly in data-driven approaches, can lead to very promising models capable of addressing the inverse kinematic problem with much greater performance compared to existing models. 
The retrieval results can be further optimized locally using different optimization approaches, specifically gradient-based optimization is possible using our solver. Furthermore, the depth of the dataset demonstrated in the study also illustrates how this dataset can be utilized in understanding the space of possible paths in 1-DOF planar linkage mechanisms, and assessing the possibility of replicating target paths using said mechanisms.

\section*{LIMITATIONS}
A key limitation of the LINKS dataset is the type of mechanisms it includes. Currently, the dataset is limited to mechanisms with simple kinematic loops, which was mainly due to the choice of the solver. This means that the dataset in its current version does not include mechanisms with complex kinematic loops such as the double-butterfly mechanism.  However as we demonstrated in the prior section, the depth of paths covered by our dataset is still significant. Another limitation of the dataset is the fact that the simulations and shapes are provided for an entire revolution of the actuator arm which means that this dataset is limited to mechanisms that are non-locking for the entire revolution of the actuator. Therefore, partially non-locking mechanisms and partial paths are not included in the dataset. These limitations can be addressed by adjusting our current generative scheme, however, that may impact the computation time to evaluate each design.
Finally, the current version of the dataset only contains mechanisms made up of joints. Other mechanical components~(such as sliders) are not included in the current dataset and will be part of future work. Our vision of LINKS is to continually evolve it with more sophisticated components and mechanisms being added in future versions of the dataset. Specifically, simple components such as sliders and power transmission components such as gears, belts and pulleys, \etc can be added to the same system. We hope LINKS will enable researchers to explore data-driven approaches for inverse kinematic synthesis.

\section*{CONCLUSION \& FUTURE WORK}
In this paper, we introduce LINKS, the first large-scale dataset of planar linkage mechanisms with over 100 million 1-DOF planar linkage mechanisms and 1.1 billion coupler curves. 
 We also provide normalized paths resulting from all mechanisms to enable easier comparison to target paths and exploration of the output space of 1-DOF mechanisms. Furthermore, we curate the normalized paths to remove biases in the data. 
Besides the data, we also demonstrate the depth of paths covered by the dataset in a study of shape retrieval, where we show how a search for unseen random query paths can identify mechanisms that can generate them. This shows that the current dataset covers the space of possible curves produced by 1-DOF planar linkage mechanisms well.
Besides providing evidence of an enhanced numerical atlas approach for linkage retrieval, we hope that LINKS will lead to greater interest in the linkage synthesis problem and enable and enhance data-driven and optimization approaches for inverse kinematic problems.

\bibliographystyle{iclr2021_conference}

\section*{acknowledgment}
The authors acknowledge the MIT SuperCloud and Lincoln Laboratory Supercomputing Center for providing HPC resources that have contributed to the research results reported in this paper. We also thank MathWorks for supporting Amin Heyrani Nobari's studies while working on this project. 
Finally, we would like to thank Dr. Wei Chen for his guidance in the initial stages of our work.

%
% ...

\bibliography{asme2e}

\begin{thebibliography}{36}
\providecommand{\natexlab}[1]{#1}
\providecommand{\url}[1]{\texttt{#1}}
\expandafter\ifx\csname urlstyle\endcsname\relax
  \providecommand{\doi}[1]{doi: #1}\else
  \providecommand{\doi}{doi: \begingroup \urlstyle{rm}\Url}\fi

\bibitem[10.(2010)]{10.1115/DETC2010-29028}
\emph{{On the Extension of a Fourier Descriptor Based Method for Four-Bar
  Linkage Synthesis for Generation of Open and Closed Paths}}, volume Volume 2:
  34th Annual Mechanisms and Robotics Conference, Parts A and B of
  \emph{International Design Engineering Technical Conferences and Computers
  and Information in Engineering Conference}, 08 2010.
\newblock \doi{10.1115/DETC2010-29028}.
\newblock URL \url{https://doi.org/10.1115/DETC2010-29028}.

\bibitem[10.(2018)]{10.1115/DETC2018-85529}
\emph{{Kinematic Synthesis Using Reinforcement Learning}}, volume Volume 2A:
  44th Design Automation Conference of \emph{International Design Engineering
  Technical Conferences and Computers and Information in Engineering
  Conference}, 08 2018.
\newblock \doi{10.1115/DETC2018-85529}.
\newblock URL \url{https://doi.org/10.1115/DETC2018-85529}.
\newblock V02AT03A009.

\bibitem[B\"{a}cher et~al.(2015)B\"{a}cher, Coros, and
  Thomaszewski]{bacher_2015}
Moritz B\"{a}cher, Stelian Coros, and Bernhard Thomaszewski.
\newblock Linkedit: Interactive linkage editing using symbolic kinematics.
\newblock \emph{ACM Trans. Graph.}, 34\penalty0 (4), jul 2015.
\newblock ISSN 0730-0301.
\newblock \doi{10.1145/2766985}.
\newblock URL \url{https://doi-org.libproxy.mit.edu/10.1145/2766985}.

\bibitem[Brown et~al.(2020)Brown, Mann, Ryder, Subbiah, Kaplan, Dhariwal,
  Neelakantan, Shyam, Sastry, Askell, Agarwal, Herbert-Voss, Krueger, Henighan,
  Child, Ramesh, Ziegler, Wu, Winter, Hesse, Chen, Sigler, Litwin, Gray, Chess,
  Clark, Berner, McCandlish, Radford, Sutskever, and Amodei]{brown2020language}
Tom~B. Brown, Benjamin Mann, Nick Ryder, Melanie Subbiah, Jared Kaplan,
  Prafulla Dhariwal, Arvind Neelakantan, Pranav Shyam, Girish Sastry, Amanda
  Askell, Sandhini Agarwal, Ariel Herbert-Voss, Gretchen Krueger, Tom Henighan,
  Rewon Child, Aditya Ramesh, Daniel~M. Ziegler, Jeffrey Wu, Clemens Winter,
  Christopher Hesse, Mark Chen, Eric Sigler, Mateusz Litwin, Scott Gray,
  Benjamin Chess, Jack Clark, Christopher Berner, Sam McCandlish, Alec Radford,
  Ilya Sutskever, and Dario Amodei.
\newblock Language models are few-shot learners, 2020.

\bibitem[Cabrera et~al.(2002)Cabrera, Simon, and Prado]{CABRERA20021165}
J.A. Cabrera, A.~Simon, and M.~Prado.
\newblock Optimal synthesis of mechanisms with genetic algorithms.
\newblock \emph{Mechanism and Machine Theory}, 37\penalty0 (10):\penalty0
  1165--1177, 2002.
\newblock ISSN 0094-114X.
\newblock \doi{https://doi.org/10.1016/S0094-114X(02)00051-4}.
\newblock URL
  \url{https://www.sciencedirect.com/science/article/pii/S0094114X02000514}.

\bibitem[Chu \& Sun(2010{\natexlab{a}})Chu and Sun]{10.1115/1.4001774}
Jinkui Chu and Jianwei Sun.
\newblock {A New Approach to Dimension Synthesis of Spatial Four-Bar Linkage
  Through Numerical Atlas Method}.
\newblock \emph{Journal of Mechanisms and Robotics}, 2\penalty0 (4), 08
  2010{\natexlab{a}}.
\newblock ISSN 1942-4302.
\newblock \doi{10.1115/1.4001774}.
\newblock URL \url{https://doi.org/10.1115/1.4001774}.
\newblock 041004.

\bibitem[Chu \& Sun(2010{\natexlab{b}})Chu and Sun]{CHU2010867}
Jinkui Chu and Jianwei Sun.
\newblock Numerical atlas method for path generation of spherical four-bar
  mechanism.
\newblock \emph{Mechanism and Machine Theory}, 45\penalty0 (6):\penalty0
  867--879, 2010{\natexlab{b}}.
\newblock ISSN 0094-114X.
\newblock \doi{https://doi.org/10.1016/j.mechmachtheory.2009.12.005}.
\newblock URL
  \url{https://www.sciencedirect.com/science/article/pii/S0094114X09002286}.

\bibitem[Ciresan et~al.(2011)Ciresan, Meier, Masci, Gambardella, and
  Schmidhuber]{mnist}
Dan~C. Ciresan, Ueli Meier, Jonathan Masci, Luca~Maria Gambardella, and
  J{\"{u}}rgen Schmidhuber.
\newblock High-performance neural networks for visual object classification.
\newblock \emph{CoRR}, abs/1102.0183, 2011.
\newblock URL \url{http://arxiv.org/abs/1102.0183}.

\bibitem[Deng et~al.(2009)Deng, Dong, Socher, Li, Li, and Fei-Fei]{imagenet}
Jia Deng, Wei Dong, Richard Socher, Li-Jia Li, Kai Li, and Li~Fei-Fei.
\newblock Imagenet: A large-scale hierarchical image database.
\newblock In \emph{2009 IEEE Conference on Computer Vision and Pattern
  Recognition}, pp.\  248--255, 2009.
\newblock \doi{10.1109/CVPR.2009.5206848}.

\bibitem[Deshpande \& Purwar(2019{\natexlab{a}})Deshpande and
  Purwar]{10.1115/1.4042325}
Shrinath Deshpande and Anurag Purwar.
\newblock {A Machine Learning Approach to Kinematic Synthesis of Defect-Free
  Planar Four-Bar Linkages}.
\newblock \emph{Journal of Computing and Information Science in Engineering},
  19\penalty0 (2), 02 2019{\natexlab{a}}.
\newblock ISSN 1530-9827.
\newblock \doi{10.1115/1.4042325}.
\newblock URL \url{https://doi.org/10.1115/1.4042325}.
\newblock 021004.

\bibitem[Deshpande \& Purwar(2019{\natexlab{b}})Deshpande and Purwar]{ml2}
Shrinath Deshpande and Anurag Purwar.
\newblock {Computational Creativity Via Assisted Variational Synthesis of
  Mechanisms Using Deep Generative Models}.
\newblock \emph{Journal of Mechanical Design}, 141\penalty0 (12), 09
  2019{\natexlab{b}}.
\newblock ISSN 1050-0472.
\newblock \doi{10.1115/1.4044396}.
\newblock URL \url{https://doi.org/10.1115/1.4044396}.
\newblock 121402.

\bibitem[Deshpande \& Purwar(2020)Deshpande and Purwar]{10.1115/1.4048422}
Shrinath Deshpande and Anurag Purwar.
\newblock {An Image-Based Approach to Variational Path Synthesis of Linkages}.
\newblock \emph{Journal of Computing and Information Science in Engineering},
  21\penalty0 (2), 10 2020.
\newblock ISSN 1530-9827.
\newblock \doi{10.1115/1.4048422}.
\newblock URL \url{https://doi.org/10.1115/1.4048422}.
\newblock 021005.

\bibitem[Deshpande \& Purwar(2021)Deshpande and Purwar]{deshpande2021image}
Shrinath Deshpande and Anurag Purwar.
\newblock An image-based approach to variational path synthesis of linkages.
\newblock \emph{Journal of Computing and Information Science in Engineering},
  21\penalty0 (2), 2021.

\bibitem[Ebrahimi \& Payvandy(2015)Ebrahimi and Payvandy]{EBRAHIMI2015189}
Saeed Ebrahimi and Pedram Payvandy.
\newblock Efficient constrained synthesis of path generating four-bar
  mechanisms based on the heuristic optimization algorithms.
\newblock \emph{Mechanism and Machine Theory}, 85:\penalty0 189--204, 2015.
\newblock ISSN 0094-114X.
\newblock \doi{https://doi.org/10.1016/j.mechmachtheory.2014.11.021}.
\newblock URL
  \url{https://www.sciencedirect.com/science/article/pii/S0094114X14003036}.

\bibitem[Erdman \& Sandor(1984)Erdman and Sandor]{erdman1984advanced}
A.G. Erdman and G.N. Sandor.
\newblock \emph{Advanced Mechanism Design: Analysis and Synthesis, Volume 2}.
\newblock Prentice-Hall, 1984.
\newblock URL \url{https://books.google.com/books?id=diXAswEACAAJ}.

\bibitem[Khan et~al.(2015{\natexlab{a}})Khan, Ullah, and
  Al-Grafi]{khan2015dimensional}
N~Khan, I~Ullah, and M~Al-Grafi.
\newblock Dimensional synthesis of mechanical linkages using artificial neural
  networks and fourier descriptors.
\newblock \emph{Mechanical Sciences}, 6\penalty0 (1):\penalty0 29--34,
  2015{\natexlab{a}}.

\bibitem[Khan et~al.(2015{\natexlab{b}})Khan, Ullah, and
  Al-Grafi]{ms-6-29-2015}
N.~Khan, I.~Ullah, and M.~Al-Grafi.
\newblock Dimensional synthesis of mechanical linkages using artificial neural
  networks and fourier descriptors.
\newblock \emph{Mechanical Sciences}, 6\penalty0 (1):\penalty0 29--34,
  2015{\natexlab{b}}.
\newblock \doi{10.5194/ms-6-29-2015}.
\newblock URL \url{https://ms.copernicus.org/articles/6/29/2015/}.

\bibitem[Kingma \& Welling(2014)Kingma and Welling]{kingma2014autoencoding}
Diederik~P Kingma and Max Welling.
\newblock Auto-encoding variational bayes, 2014.

\bibitem[Krizhevsky(2009)]{cifar}
Alex Krizhevsky.
\newblock Learning multiple layers of features from tiny images.
\newblock pp.\  32--33, 2009.
\newblock URL
  \url{https://www.cs.toronto.edu/~kriz/learning-features-2009-TR.pdf}.

\bibitem[Lipson(2005)]{10.1115/1.2198255}
Hod Lipson.
\newblock {A Relaxation Method for Simulating the Kinematics of Compound
  Nonlinear Mechanisms}.
\newblock \emph{Journal of Mechanical Design}, 128\penalty0 (4):\penalty0
  719--728, 10 2005.
\newblock ISSN 1050-0472.
\newblock \doi{10.1115/1.2198255}.
\newblock URL \url{https://doi.org/10.1115/1.2198255}.

\bibitem[Lipson(2008)]{lipson_2008}
Hod Lipson.
\newblock Evolutionary synthesis of kinematic mechanisms.
\newblock \emph{Artificial Intelligence for Engineering Design, Analysis and
  Manufacturing}, 22\penalty0 (3):\penalty0 195–205, 2008.
\newblock \doi{10.1017/S0890060408000139}.

\bibitem[McCarthy \& Soh(2010)McCarthy and Soh]{mccarthy2010geometric}
J~Michael McCarthy and Gim~Song Soh.
\newblock \emph{Geometric design of linkages}, volume~11.
\newblock Springer Science \& Business Media, 2010.

\bibitem[McGarva(1994)]{MCGARVA1994223}
John~R McGarva.
\newblock Rapid search and selection of path generating mechanisms from a
  library.
\newblock \emph{Mechanism and Machine Theory}, 29\penalty0 (2):\penalty0
  223--235, 1994.
\newblock ISSN 0094-114X.
\newblock \doi{https://doi.org/10.1016/0094-114X(94)90032-9}.
\newblock URL
  \url{https://www.sciencedirect.com/science/article/pii/0094114X94900329}.

\bibitem[Mnih et~al.(2013)Mnih, Kavukcuoglu, Silver, Graves, Antonoglou,
  Wierstra, and Riedmiller]{mnih2013playing}
Volodymyr Mnih, Koray Kavukcuoglu, David Silver, Alex Graves, Ioannis
  Antonoglou, Daan Wierstra, and Martin Riedmiller.
\newblock Playing atari with deep reinforcement learning, 2013.

\bibitem[Radhakrishnan \& Campbell(2011)Radhakrishnan and Campbell]{campbell}
Pradeep Radhakrishnan and Matthew~I. Campbell.
\newblock A graph grammar based scheme for generating and evaluating planar
  mechanisms.
\newblock In John~S. Gero (ed.), \emph{Design Computing and Cognition '10},
  pp.\  663--679, Dordrecht, 2011. Springer Netherlands.
\newblock ISBN 978-94-007-0510-4.

\bibitem[Regenwetter et~al.(2021)Regenwetter, Nobari, and
  Ahmed]{DBLP:journals/corr/abs-2110-10863}
Lyle Regenwetter, Amin~Heyrani Nobari, and Faez Ahmed.
\newblock Deep generative models in engineering design: {A} review.
\newblock \emph{CoRR}, abs/2110.10863, 2021.
\newblock URL \url{https://arxiv.org/abs/2110.10863}.

\bibitem[Reuleaux(1875)]{reuleaux1875lehrbuch}
Franz Reuleaux.
\newblock \emph{Lehrbuch der Kinematik}, volume~1.
\newblock Vieweg, 1875.

\bibitem[Sharma \& Purwar(2020)Sharma and Purwar]{10.1115/1.4046817}
Shashank Sharma and Anurag Purwar.
\newblock {Using a Point-Line-Plane Representation for Unified Simulation of
  Planar and Spherical Mechanisms}.
\newblock \emph{Journal of Computing and Information Science in Engineering},
  20\penalty0 (6), 05 2020.
\newblock ISSN 1530-9827.
\newblock \doi{10.1115/1.4046817}.
\newblock URL \url{https://doi.org/10.1115/1.4046817}.
\newblock 061002.

\bibitem[Sohn et~al.(2015)Sohn, Lee, and Yan]{NIPS2015_8d55a249}
Kihyuk Sohn, Honglak Lee, and Xinchen Yan.
\newblock Learning structured output representation using deep conditional
  generative models.
\newblock In C.~Cortes, N.~Lawrence, D.~Lee, M.~Sugiyama, and R.~Garnett
  (eds.), \emph{Advances in Neural Information Processing Systems}, volume~28.
  Curran Associates, Inc., 2015.
\newblock URL
  \url{https://proceedings.neurips.cc/paper/2015/file/8d55a249e6baa5c06772297520da2051-Paper.pdf}.

\bibitem[Suh \& Radcliffe(1987)Suh and Radcliffe]{suh_radcliffe_1987}
Chung~Ha Suh and Charles~W. Radcliffe.
\newblock \emph{Kinematics and mechanisms design}.
\newblock Robert E. Krieger, 1987.

\bibitem[Sun et~al.(2015)Sun, Lu, and Chu]{doi:10.1080/17415977.2014.890615}
Jianwei Sun, He~Lu, and Jinkui Chu.
\newblock Variable step-size numerical atlas method for path generation of
  spherical four-bar crank-slider mechanism.
\newblock \emph{Inverse Problems in Science and Engineering}, 23\penalty0
  (2):\penalty0 256--276, 2015.
\newblock \doi{10.1080/17415977.2014.890615}.
\newblock URL \url{https://doi.org/10.1080/17415977.2014.890615}.

\bibitem[Ullah \& Kota(1997)Ullah and Kota]{10.1115/1.2826396}
Irfan Ullah and Sridhar Kota.
\newblock {Optimal Synthesis of Mechanisms for Path Generation Using Fourier
  Descriptors and Global Search Methods}.
\newblock \emph{Journal of Mechanical Design}, 119\penalty0 (4):\penalty0
  504--510, 12 1997.
\newblock ISSN 1050-0472.
\newblock \doi{10.1115/1.2826396}.
\newblock URL \url{https://doi.org/10.1115/1.2826396}.

\bibitem[Varedi-Koulaei \& Rezagholizadeh(2020)Varedi-Koulaei and
  Rezagholizadeh]{doi:10.1177/0954406220908616}
SM~Varedi-Koulaei and H~Rezagholizadeh.
\newblock Synthesis of the four-bar linkage as path generation by choosing the
  shape of the connecting rod.
\newblock \emph{Proceedings of the Institution of Mechanical Engineers, Part C:
  Journal of Mechanical Engineering Science}, 234\penalty0 (13):\penalty0
  2643--2652, 2020.
\newblock \doi{10.1177/0954406220908616}.
\newblock URL \url{https://doi.org/10.1177/0954406220908616}.

\bibitem[Vasiliu \& Yannou(2001)Vasiliu and Yannou]{ml1}
Adrian Vasiliu and Bernard Yannou.
\newblock Dimensional synthesis of planar mechanisms using neural networks:
  Application to path generator linkages.
\newblock \emph{Mechanism and Machine Theory}, 36:\penalty0 299--310, 02 2001.
\newblock \doi{10.1016/S0094-114X(00)00037-9}.

\bibitem[Waldron \& Sreenivasan(1996)Waldron and
  Sreenivasan]{10.1115/1.2826898}
K.~J. Waldron and S.~V. Sreenivasan.
\newblock {A Study of the Solvability of the Position Problem for Multi-Circuit
  Mechanisms by Way of Example of the Double Butterfly Linkage}.
\newblock \emph{Journal of Mechanical Design}, 118\penalty0 (3):\penalty0
  390--395, 09 1996.
\newblock ISSN 1050-0472.
\newblock \doi{10.1115/1.2826898}.
\newblock URL \url{https://doi.org/10.1115/1.2826898}.

\bibitem[Wu et~al.(2011)Wu, Ge, Gao, and Guo]{10.1115/1.4004227}
Jun Wu, Q.~J. Ge, Feng Gao, and W.~Z. Guo.
\newblock {On the Extension of a Fourier Descriptor Based Method for Planar
  Four-Bar Linkage Synthesis for Generation of Open and Closed Paths}.
\newblock \emph{Journal of Mechanisms and Robotics}, 3\penalty0 (3), 07 2011.
\newblock ISSN 1942-4302.
\newblock \doi{10.1115/1.4004227}.
\newblock URL \url{https://doi.org/10.1115/1.4004227}.
\newblock 031002.

\end{thebibliography}



\end{document}